\documentclass{article}

\usepackage{amsmath,amsfonts,amssymb,amsthm}
\usepackage{graphicx}           % Needed for including graphics.
\usepackage{url}                % Facility for activating URLs.
\usepackage[colorlinks = true, linkcolor = red, urlcolor = red, citecolor = red, anchorcolor = red]{hyperref}
\usepackage[ruled,vlined]{algorithm2e}
\usepackage{subcaption}

\usepackage[a4paper]{geometry}

\theoremstyle{definition}

\begin{document}

\title{DP-XGBoost: Private Machine Learning at Scale}
\author{Nicolas Grislain\\
\url{ng@sarus.tech} \\ 
\and 
Joan Gonzalvez \\
\url{jg@sarus.tech}}
\date{}
\maketitle

\begin{abstract}
The big-data revolution announced ten years ago \cite{manyika2011big}
does not seem to have fully happened at the expected scale \cite{analytics2016age}.
One of the main obstacle to this, has been the lack of data circulation.
And one of the many reasons people and organizations did not share as much as expected
is the \emph{privacy risk} associated with data sharing operations.

There has been many works on practical systems to compute statistical
queries with \emph{Differential Privacy} (DP). There have also been practical implementations of
systems to train Neural Networks with DP \cite{mcmahan2018general, opacus}, but relatively little efforts
have been dedicated to designing scalable classical \emph{Machine Learning} (ML) models providing
DP guarantees.

In this work we describe and implement a DP fork of a battle tested ML model: \emph{XGBoost}.
Our approach beats by a large margin previous attempts at the task, in terms of accuracy achieved for a given privacy budget.
It is also the only DP implementation of boosted trees that scales to big data and can run in
distributed environments such as: \href{https://kubernetes.io/}{Kubernetes}, \href{https://dask.org/}{Dask} or \href{https://spark.apache.org/}{Apache Spark}.
\end{abstract}

% Introduction and motivation 
\section{Introduction}

As more and more data is collected on individuals and as data science techniques become more powerful,
threats to privacy have multiplied and strong concerns have emerged.
Not only do these privacy risks pose a threat to civic liberties, but they also hamper innovation in industries
such as \emph{health care}, \emph{personal finance} or \emph{smart cities}, because the necessity to protect
privacy is not always compatible with the current requirements of data science.

Recently, \emph{differential privacy} (DP) \cite{dwork2014} has emerged as the \emph{gold standard} of
privacy protection. Indeed, differential privacy provides a theoretical framework for measuring and bounding
the \emph{privacy loss} occuring when one publishes an analysis on a private dataset.
It provides a formal protection while not requiring too many conditions.
It also comes with many nice building blocks and composition theorems enabling many applications in
data analysis, from basic statistics to advanced \emph{artificial intelligence} algorithms.

\subsection{Private data analysis}

Many research papers develop the theoretical framework of differential privacy as referred by
the monography \cite{dwork2014} from Dwork and Roth. Surprisingly, few practical implementations
and applications have been proposed.

In the field of analytics, McSherry proposed in 2009 a system: \emph{PINQ}, to run SQL-like queries
with differential privacy guarantees \cite{mcsherry2009privacy}, but this system does not suport
one-to-many \verb"Joins" due to the difficulty to bound the contribution of a record.
Proserpio et al. \cite{proserpio2012calibrating} propose a version for weighted dataset alleviating
this restriction.
Based on an extensive review of existing SQL practices and systems, Johnson et alii \cite{johnson2018towards}
propose a new approach to private SQL, leveraging a variant of \emph{smooth sensitivity} \cite{nissim2007smooth}.
Kotsogiannis et alii \cite{kotsogiannis2019privatesql, kotsogiannis2019architecting} designed a system that allows
relational data with foreign keys and where queries are executed on private synopses to limit their consumption.
Wilson et alii also introduced a system \cite{wilson2019differentially} where they focus on bounding user contribution
in the case many rows relate to the same individual.
To save privacy \cite{kotsogiannis2019privatesql, kotsogiannis2019architecting} designed a system to run SQL queries
on private synopses of the data.
Recently a team at Google proposed \footnote{An implementation is available there: \href{https://github.com/google/differential-privacy}{https://github.com/google/differential-privacy}}
a DP version of SQL \cite{wilson2019differentially} with a focus on bounding user contribution even when it appears more than once.

Beyond simple statistics and SQL-like analytics, many efforts have been devoted to the developpement of
private deep-learning and generic \emph{Differentially Private Stochastic Gradient Descent} (DP-SGD)
\cite{abadi2016deep, papernot2018scalable, kairouz2021practical}, with some practical successes \cite{mcmahan2017learning}.
In the same field, ways of accounting for privacy-loss have been improved with DP-SGD in mind \cite{mironov2017renyi, bu2020deep}.
The most famous implementations of DP-SGD are \href{https://github.com/tensorflow/privacy}{Tensorflow Privacy} \cite{mcmahan2018general}
and \href{https://github.com/pytorch/opacus}{Opacus} \cite{opacus}.

There has been relatively little efforts to bring differential privacy to classical machine learning in practice;
appart from generic convex optimization methods \cite{wu2017bolt, iyengar2019towards} and a few implementations
designed mostly for research and not for scalability such as \cite{diffprivlib}.

The goal of this work is to enable state-of-the-art machine learning with differential privacy at scale.
The scalability is particularily important since the larger the number of individuals in the data the cheapest privacy is
in terms of accuracy. We decided to design and build a private boosted tree solution, because of the popularity in the industry of
regression tree based methods \cite{wu2008top} and boosted trees in particular (see:
\href{https://storage.googleapis.com/kaggle-media/surveys/Kaggle's\%20State\%20of\%20Machine\%20Learning\%20and\%20Data\%20Science\%202021.pdf}{Kaggle - State of Machine Learning and Data Science 2021}).
To build a scalable solution, we started from a mature platform.
There are many implementations of regression trees, among which XGBoost \cite{chen2016xgboost},
LightGBM \cite{ke2017lightgbm} or CatBoost \cite{prokhorenkova2017catboost}.
Because of its adoption and scalability, XGBoost was a good foundation to build upon.

\section{Definition and properties of differential privacy}
% Section on differential privacy def + exp mech + laplace + composition

We consider a dataset $\mathbf{D} = (x_1,\dots,x_n)\in \mathcal{X}^n$, where $\mathcal{X}$ is the feature space and $n\geq 1$ is the sample size. 
Two datasets $D$ and $D'$ are said to be \textit{neighboring} if they differ by one single instance. Differential privacy aims at controlling the probability that a 
single sample modifies the output of a real function or query $f(D) \in \mathbb{R}$ significantly. Let $\epsilon > 0$, a random function $f$ is called $\epsilon$-differentially private 
($\epsilon$-DP) if the following condition is verified. For any two neighboring datasets $D$ and $D'$ and output set $O \subset \mathbb{R}$  we have:

\begin{equation*}
  \mathbb{P}\left( f(D) \in O \right) \leq e^{\epsilon} \mathbf{P}\left( f(D') \in O \right) 
\end{equation*}

We'll be particularily interested in two $\epsilon$-DP mechanisms: the Laplace Mechanism and the Exponential Mechanism (\cite{dwork2014}, \cite{mcsherry2007mechanism}). The Laplace Mechanism 
allows one to release a private query $f(D)$ by adding a Laplace distributed noise. 

Define the \textit{sensitivity} of a query $f$ by: 

\begin{equation*}
  \Delta f = \max_{D, D'} \vert f(D) - f(D') \vert 
\end{equation*}

where the maximum is taken over all neighboring datasets. Then the following mechanism is $\epsilon$-DP \cite{dwork2014}:

\begin{equation*}
  f(D) + \text{Lap}\left( \frac{\Delta f}{\epsilon} \right) 
\end{equation*}

Another well studied $\epsilon$-DP mechanism is the Exponential Mechanism \cite{mcsherry2007mechanism} which allows the differentially private 
selection of the best candidate from a given list and utility function $u(D, x)$. If the output is distributed as  

\begin{equation*} 
\mathbb{P}(f(D) = x) \propto e^{ \frac{\epsilon u(D,x)}{\Delta u} }
\end{equation*}

then the Exponential selection mechanism is $\epsilon$-DP. 

When performing several $\epsilon$-DP queries one can be interested in the overall differential privacy 
of the resulting \textit{composed} query. We'll make use of the two following \textit{composition theorems} (\cite{dwork2014}). 
Let $f_1, \cdots, f_n$ be a series of $\epsilon_1, \cdots, \epsilon_n$-DP queries. If the queries are applied sequentially (\textit{sequential composition}) on the dataset, the resulting mechanism 
is $\sum_{i=1}^{n} \epsilon_i$-DP. Instead if the $f_i$ are applied on disjoint subsets of the dataset $D$ (\textit{parallel composition}), then the overall composed mechanism is only $\max_{1 \leq i \leq n} \epsilon_i$.

\section{Differentially Private Gradient Boosted Trees} 

Given a dataset $\mathcal{D} = \left\{ (x_i, y_i ) \right\}$ with $n$ training instances 
$x_i \in \mathbb{R}^m$ and labels $y_i \in [-1, 1]$, gradient boosted trees aim at learning an additive
model consisting of $K$ classification or regression trees (CART) $f_k$, so that the prediction for a sample $x$ is given by:

\begin{equation*} 
f(x) = \sum_{k = 1}^{K} f_k(x) 
\end{equation*}

The learning objective in XGBoost is a regularized loss:

\begin{equation} 
\mathcal{L} = \sum_{i=1}^n l( f(x_i), y_i) +  \sum_{k=1}^{K} \Omega(f_k) 
\end{equation} 

Where $\Omega(f_k)$ is the regularization term for the $k$-th tree, which is defined as the sum of squares of leaf values of the tree. $l(.,.)$ is a loss function which can correspond to classification or regression problems, \textsl{e.g.} $l(
\hat{y}, y) = \frac 12 (\hat{y} - y)^2$, which we will choose for simplicity and turning classification problems into regression ones for our numerical experiments. 

To solve this optimization problem the usual approach is first to use a greedy method. For the $k$-th tree, the objective will be to minimize the following:

\begin{equation*}
\mathcal{L} = \sum_{i=1}^n l(y_i, \hat{y}_i^{k-1} + f_k(x_i)) +   \Omega(f_k) 
\end{equation*}

Where $\hat{y}_i^{k-1}$ is the prediction of the model using already constructed trees: $\hat{y}_i^{k-1} = \sum_{j=1}^{k-1} f_j(x_i)$. In order to obtain a simple and quick tree building algorithm we then minimize the second-order approximation of the greedy objective:

\begin{equation} 
\sum_{i=1}^n  g_i f_k(x_i) + \frac 12 h_i f_k(x_i)^2 + \Omega(f_k) 
\end{equation} 

with $g_i$ and $h_i$ being respectively the first and second-order derivative of the loss function $l$ with respect to the previous model output $\hat{y}_i^{k-1}$. This objective allows one to find an exact greedy tree-building algorithm. Recall that in a CART, each node splits the input space depending on whether the feature at some index value is greater than a threshold. Thus, associated to a tree structure with leaves instances $I_j$ (the set of training samples which end up in leaf $j$), we can show that the minimal objective is given by: 

\begin{equation} 
- \frac 12 \sum_{j=1}^{T} \frac{\left( \sum_{i \in I_j} g_i \right)^2}{\sum_{i \in I_j} h_i + \lambda}  
\label{objective}
\end{equation} 

Where $T$ is the number of leaves. This score is used to construct a single tree in a greedy recursive approach: a node is expanded by choosing the split that maximizes the score reduction between the parent node and the two candidate left and right instance sets $I_L$ and $I_R$. In the case of regression the gain from splitting a node into instance $I_L$ and $I_R$ is thus given by:
\begin{equation*} 
G(I_L, I_R) = \frac{ \left( \sum_{i \in I_L} g_i \right)^2}{|I_L| + \lambda} + \frac{ \left( \sum_{i \in I_R} g_i \right)^2}{|I_R| + \lambda}
\end{equation*} 

The basic exact algorithm consists in selecting the split, among all possible feature values and indices, that maximizes this gain and typically stop when the maximum height is reached, if the maximal gain is too small, or if the number of samples in the resulting children is too small. The final nodes (leaves) are then assigned the optimal value which will constitute the tree prediction:
\begin{equation}
V(I) = - \frac{\sum_{i \in I} g_i}{\vert I \vert + \lambda}
\label{leafValue}
\end{equation}

\subsection{XGBoost}

The basic exact algorithm described above is not practical for large datasets as it requires looping over every possible split of every (possibly continuous) feature.
XGBoost aims at providing a large-scale gradient boosted trees library and solving this problem. It is written in C++ and presents several features which make
it able to handle datasets with billions of examples. It comes with interfaces in Python, Java, and Spark (in Scala) which can be used for distributed learning.

One of these optimizations is using a compressed column (CSC) format to store the dataset,
as a lot of steps (computing quantiles, finding the best splits) are faster when the feature values are stored in order.
The dataset is split into \textit{block} that fit in memory and can be distributed accross machines. 

XGBoost is also able to handle sparsity and missing values by enumerating only the non-missing or non-zero splits and computing the gain twice:
sending the default values to the left or the right node. This improves the computation drastically for the sparse datasets and the default direction is thus learnt from data.

The global approximate algorithm relies on the following idea: instead of finding the optimal splits among all possible features values,
XGBoost first computes proposal splits based on percentiles of data. In order to do that, it uses a quantile sketching algorithm, which constructs a \textit{quantile summary} 
for each partition (\textit{block}) of the training set, and allows for the efficient computation of approximate percentiles of each feature even for a large scale dataset.
The candidates are either recomputed at every level of the tree, or once per tree (the global approximate method).
\cite{chen2016xgboost} show that using the global method with an order of magnitude of 50 proposal splits per feature actually gives a very good trade-off between accuracy and tree-building time.
To implement differential privacy into XGBoost we will thus use this method. 

\section{Differentially Private Boosted Trees} 

While \cite{li2020privacy} provide a differentially private version of gradient boosted trees, they only consider the exact tree construction algorithm,
which loops through every feature value and is not scalable and slow for modern datasets. Our objective is thus to add differential privacy to the XGBoost library,
leveraging its large-scale training capabilities. 

In order to add differential privacy to XGBoost we distinguish three steps of the approximate learning algorithm. First, the 
construction of "quantile sketchs" for each feature as described previously. Second, the split selection based on a histogram
constructed using the final quantile sketchs for each feature. Finally, the leaf value calculation which can be thought of as computing
the average gradient of data samples belonging to the leaf. The mechanisms are then composed sequentially to produce an overall differentially-private tree building algorithm.

We use here the gradient data filtering (GDF) technique introduced by \cite{li2020privacy}:
the labels are assumed to be bounded in $[-1,1]$ and a universal constant (independent of training data and depending only on the loss function)
$g_l^{*}$ filters the samples by gradient before the training of each tree.
This filtering allows to compute the exact sensitivity of mechanisms for split finding and leaf value calculations.

\subsection{Sketching} 

The weighted quantile sketching algorithm is one of the feature of XGBoost which makes it able to deal with datasets containing millions of samples.
While quantile sketching algorithms existed before \cite{chen2016xgboost}, the weighted quantile sketch described in their Appendix allows one to construct a histogram by weight rather than by count.
We will briefly remind the main ideas of quantile sketching in XGBoost.

Note that equation \ref{objective} can be interpreted as a weighted least squares loss with a weight for instance $i$ given by $h_i$, the Hessian.
Thus XGBoost aims at finding for each feature $k$ a set of candidates $\left\{ s_{k,1}, \cdots, s_{k,l}\right\}$, in increasing order,
such that the $s_{k,j}$ are evenly distributed by weight, that is that the sum of Hessian of instances in the range $[s_{k,j}, s_{k,j+1}[$
is less than a parameter $\epsilon$ (the sketch accuracy). The quantile sketch builds a quantile summary which contains the minimal information necessary to compute the candidates with a given accuracy.
Two quantile summaries can be merged together, and a summary that becomes too large can be pruned down,
both these operations increase the approximation error $\epsilon$. An initial summary is constructed by starting from a subset of the data points and computing their exact ranks.
XGBoost then repeatedly merges and prunes the quantile summaries for each feature in a distributed manner (recall that the dataset is stored in a sorted format on different in-memory \textit{blocks}). 

To make this step differentially private, one could try to design a DP (weighted) quantile sketching algorithm.
This remains an open and difficult problem as even simply computing DP quantiles is not easy (see for instance the JointExp algorithm of \cite{gillenwater2021}).
The solution we adopt is to simply feed the sketch with differentially private data.
We require the knowledge of differentially private bounds (maximum and minimum) of each feature (which will be needed in any case for the DP guarantees of \cite{li2020privacy} to hold).
We then build a simple histogram of weights, with a fixed number of bins and equal spacing, for each feature and noise every bin count or weight with a Laplace query.
Points are then passed to the sketch construction algorithm of XGBoost with differentially private points and weights, making the quantile sketch a post-processing step.
Note that the histogram query for a given sketch benefits from parallel composition since the bins are by construction disjoint (\cite{dwork2014}), thus the privacy budget for each sketch can be applied to each bin. 

\begin{figure}[h]
\centering 
\includegraphics[scale=0.7]{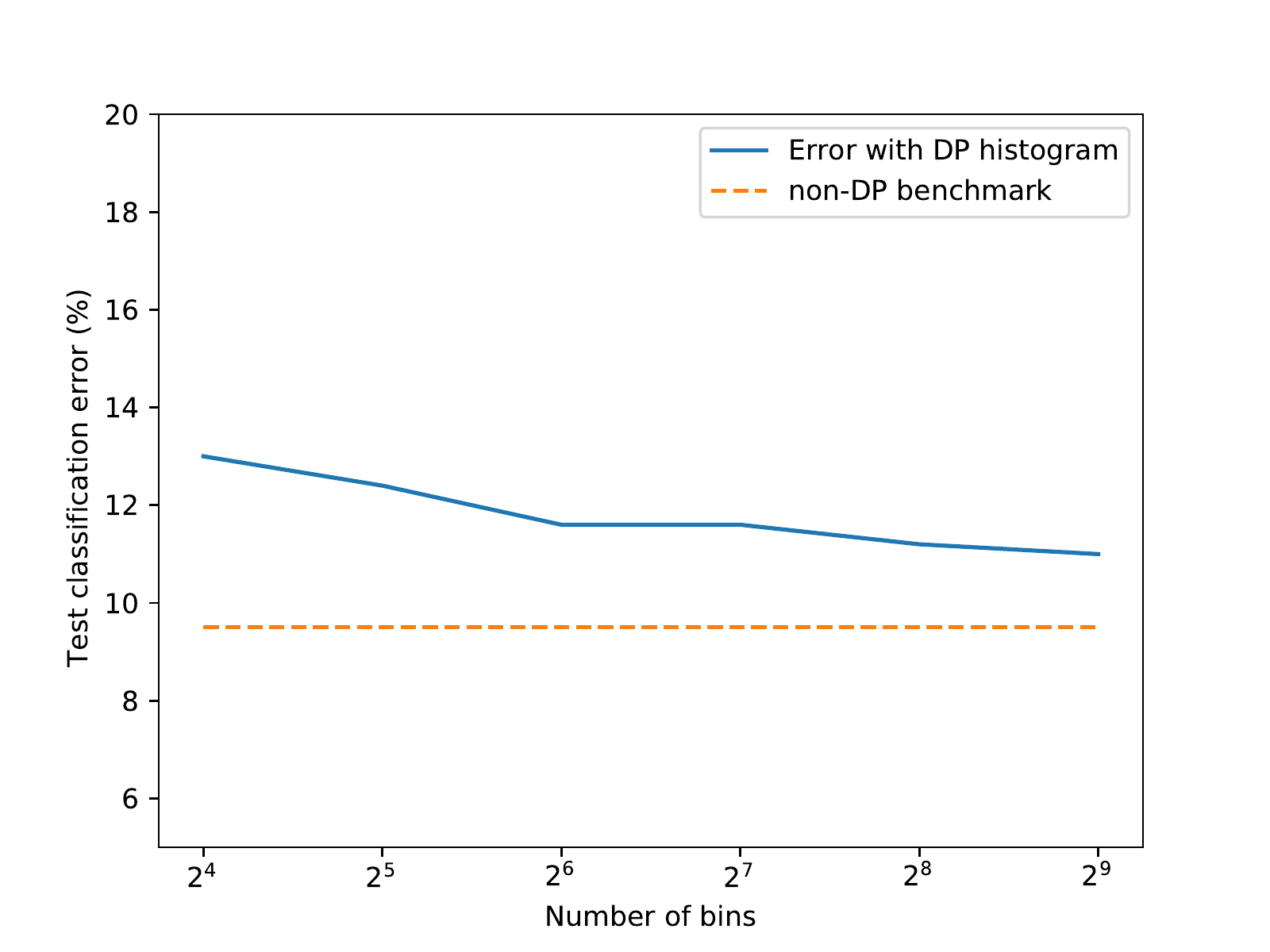}
\caption{Impact of bin size of the DP histogram to feed the weighted quantile sketching algorithm}
\label{sketchAccuracy2}
\end{figure}

The impact of this DP pre-processing is not compromising on the performance of the trees. Indeed figure~\ref{sketchAccuracy2} shows the test accuracy,
for a real-world binary classification task, using quantile sketch with this DP method versus the baseline for different number of bins and $\epsilon = 1$.
We can see the added approximation error from building the XGBoost sketchs with a noisy differentially-private Hessian histogram is marginal. 

\subsection{Split selection}

For split selection, recall Algorithm 2 from \cite{chen2016xgboost}: given weighted quantile sketches $S_k = \left\{ s_{k,1}, \cdots, s_{k,l} \right\} $ for a feature $k$ at a given node to expand we compute a histogram of gradients and Hessians for each proposed split: 
\begin{eqnarray*} 
G_{kv} &=& \sum_{j\ \text{s.t}\ s_{k,v-1} < x_{jk} \leq s_{k,v} } g_j \\
H_{kv} &=& \sum_{j\ \text{s.t}\ s_{k,v-1} < x_{jk} \leq s_{k,v} } h_j 
\end{eqnarray*}

Where the statistics are collected only for the non-missing entries as part of the sparsity-aware improvement of XGBoost. The gain is then computed twice for each proposed split using this histogram, first by sending missing values to the left child, and then sending them to the right child, as is described in Algorithm 3 from \cite{chen2016xgboost}.

Selecting the best split in a DP manner is then not more complex than in the basic exact algorithm considered in \cite{li2020privacy}, we use an exponential mechanism, described in \cite{dwork2014}: it allows one to select a candidate 
in a list where each candidate $s$ is assigned a utility function $u(X,s)$ which depends on the dataset and has a global utility of $\Delta u$. If samples the candidate from the following distribution, then the choice is guaranteed to be $\epsilon$-DP: 
\begin{equation*}
    p(s) \propto e^{ \frac{\epsilon u(X,s)}{2 \Delta u}}
\end{equation*}

For this purpose \cite{li2020privacy} compute the sensitivity of the gain defined previously thanks to the gradient filtering which allows for a universal bound on the gradients $g_i$ of the training samples. The sensitivity of the gain $\Delta G$ is given by: 
\begin{equation}
\Delta G = 3 {g_l^*}^2
\end{equation} 

Thus, the probability of sampling candidate split $j$ with a gain $G_j$ is: 

$$  p_j \propto e^{ \frac{\epsilon G_j}{6 {g_l^*}^2}}$$ 

In order to sample from this distribution without numerical errors, we use the "racing trick" of \cite{medina2020duff}. It allows one
to sample from a distribution using only log-probabilities and a single pass over the set and without having to compute the normalizing constant. 

\subsection{Leaf values} 

Once a certain criterion is met, a node is not expanded further and turned into a leaf. The criteria typically used are: reaching a maximum depth, having a maximum gain that is too small compared to a given constant, or having a sum of Hessian of samples too small (for regression this means not having enough samples in the node). The leaf is then given a value according to equation \ref{leafValue}. 

With GDF it is possible to compute the sensitivity $\Delta V$ of the leaf value with respect to the training dataset. We show here the derivation given by \cite{li2020privacy} (Appendix A), consider two adjacent datasets $I_1$ and $I_2$ differing by a single sample $x_s$ with gradient $g_s$. We'd like to find the maximum of the difference $\vert V(I_1) - V(I_2) \vert$:

\begin{eqnarray*}
\vert V(I_1) - V(I_2) \vert &=& \left\vert \frac{-\sum_{i \in I_1} g_i }{n+\lambda} + \frac{\sum_{i \in I_1} g_i + g_s}{n+1+\lambda} \right\vert \\
 &=& \left\vert \frac{ (n+\lambda)g_s - \sum_{i \in I} g_i}{(n+\lambda)(n+1+\lambda)} \right\vert  
\end{eqnarray*} 

Which reaches a maximum of (remembering that $|g_i| \leq g_l^*$)  

\begin{equation}
\frac{(2n+\lambda)g_l^*}{(n+\lambda)(n+1+\lambda)} 
\label{leafSensi}
\end{equation} 

This is then upper-bounded by $\frac{g_l^*}{1+\lambda}$ and this sensitivity is used to noise the leaves with a Laplace mechanism. However we
can drastically improve this worst-case bound with two different approaches. 

Indeed, XGBoost and other gradient boosting libraries typically offer a hyperparameter which controls the minimum Hessian weight (or number of samples) needed when expanding a node. When this parameter is set it gives Eq. \ref{leafSensi} a better bound which can drastically reduce
the variance of the Laplace noise added. If we denote by $N_{\text{min}}$ the minimum number of samples required in a child to split a node further, then we obtain the following sensitivity for the Laplace mechanisms: 

\begin{equation}
\Delta V \leq \frac{2 g_l^*}{N_{\text{min}} + 1 + \lambda} 
\end{equation} 

We found in our numerical experiments that introducing this parameter in XGBoost,
$min\_child\_weight = 10$ for small datasets of with order tens of thousands of samples,
and $min\_child\_weight = 200$ for the larger ones typically didn't decrease the accuracy of the non-DP baseline estimator
but improved our differentially-private boosted trees thanks to the variance reduction obtained in the Laplace mechanism.
Note that even without differential privacy concerns, the parameter is typically
used to reduce over-fitting as it prevents building too complex trees.  

Another approach is to use noisy average algorithms from the DP literature. Indeed equation \ref{leafValue} is just an average of gradients of data that reach the leaf node.
The problem of estimating a differentially-private average is well studied in literature, as in \cite{li2016dp}.

For instance their algorithm 2.3 can be used, requiring to noise only the sum and using the exact count.
In our case, the count is guaranteed to be non-zero as leaves cannot be empty, and the $\epsilon$-DP noisy average algorithm is the following:
\begin{algorithm}
\caption{Noisy average to compute leaf values}
\KwData{Leaf instance gradients $\left\{ g_i \right\}_{i=1, \cdots, n}$, privacy budget $\epsilon$, gradient filtering constant $g_l^*$}
$S \gets \sum_{i=1}^{n} g_i$ \; 
$V \gets \frac{ S + \text{Lap}( \frac{2g_l^*}{\epsilon} )}{n + \lambda} $ \; 

\If{$|V| > g_l^*$}
{
    $V \gets g_l^* \text{sign}(V)$ \;
}
\Return $-V$
\end{algorithm}

\begin{figure}[h]
\centering 
\includegraphics[scale=0.7]{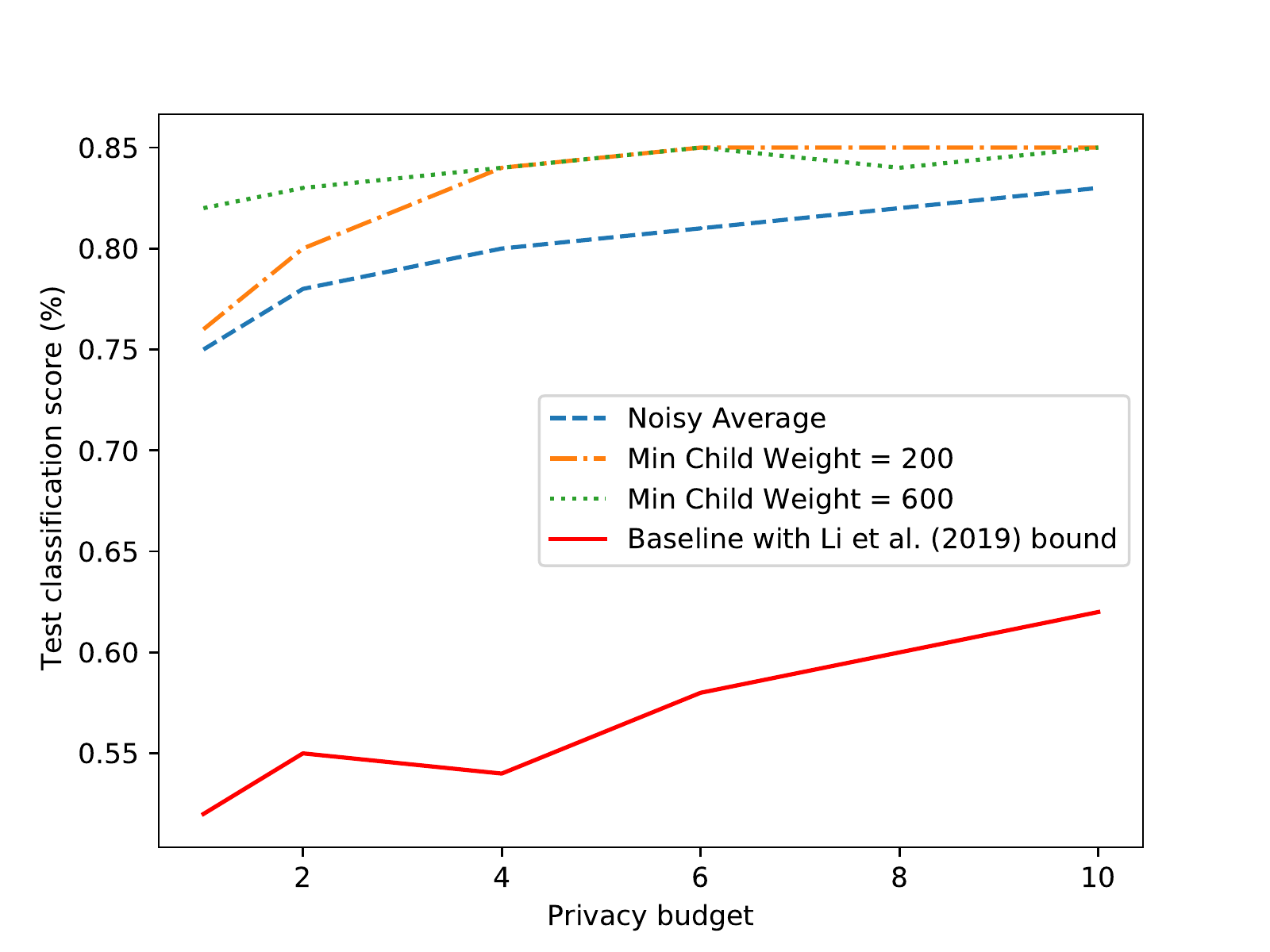}
\caption{Impact of using the minimum child weight parameter in XGBoost to reduce the leaf noise variance on the Covtype dataset}
\label{leafNoisePlot}
\end{figure}

Figure~\ref{leafNoisePlot} shows the impact of using these strategies to reduce the noise in the leaf values Laplace mechanism 
compared to using the bound given by \cite{li2020privacy}. The experiment is run on the Covtype dataset and consists in a binary
classification task with 20 trees. The baseline uses gradient filtering and geometric leaf clipping as described in \cite{li2020privacy},
while the others use the previous algorithm for NoisyAverage or a hyperparameter controlling the 
minimum number of samples needed to split a node, setting to $200$ and $600$.

\subsection{Privacy consumption and subsampling} 

Note that by design of the tree construction algorithm, the DP mechanisms benefit naturally from parallel composition:
at a given depth the dataset is split into disjoint subsets on which each mechanism acts, thus the privacy budget of a mechanism is counted once per depth level.
The privacy budget for the sketching algorithm (which is done once per tree, at the beginning of training)
is evenly divided among features and accumulated thanks to the basic composition theorem \cite{dwork2014, kairouz2015composition}.
Each quantile sketch thus receives a budget of $ \frac{ \epsilon_{\text{sketch}} }{m}$.
This budget is allocated to the histogram construction described previously and each bin count is perturbated with a Laplace noise $\text{Lap}\left( \frac{m}{\epsilon_{\text{sketch}}}\right)$

The privacy budget for a single tree $\epsilon$ is divided between exponential,
Laplace, and Histogram mechanisms in the following manner:
$\epsilon_{\text{sketch}} = \frac{\epsilon}{3}$, $\epsilon_{\text{leaf}} = \frac{\epsilon}{3}$
and $\epsilon_{\text{exponential}} = \frac{\epsilon}{3 h}$ where $h$ is the maximum depth of the tree.

We can also naturally take advantage of row subsampling in XGBoost, which as in the case of the minimum child weight,
is used to prevent overfitting and brings privacy for free. Indeed, as shown in \cite{balle2018subsampling},
if one designs an $\epsilon$-DP mechanism and runs it on a random subsampled fraction (without replacement) $0 < \gamma < 1$ of the dataset,
then the resulting mechanism has an overall exact privacy budget given by $\log\left(1 + \gamma(e^{\epsilon} - 1)\right)$), which as $\epsilon$ gets small (high-privacy regime), is of order $O(\gamma \epsilon)$. 

\begin{figure}[h]
\centering 
\includegraphics[scale=0.7]{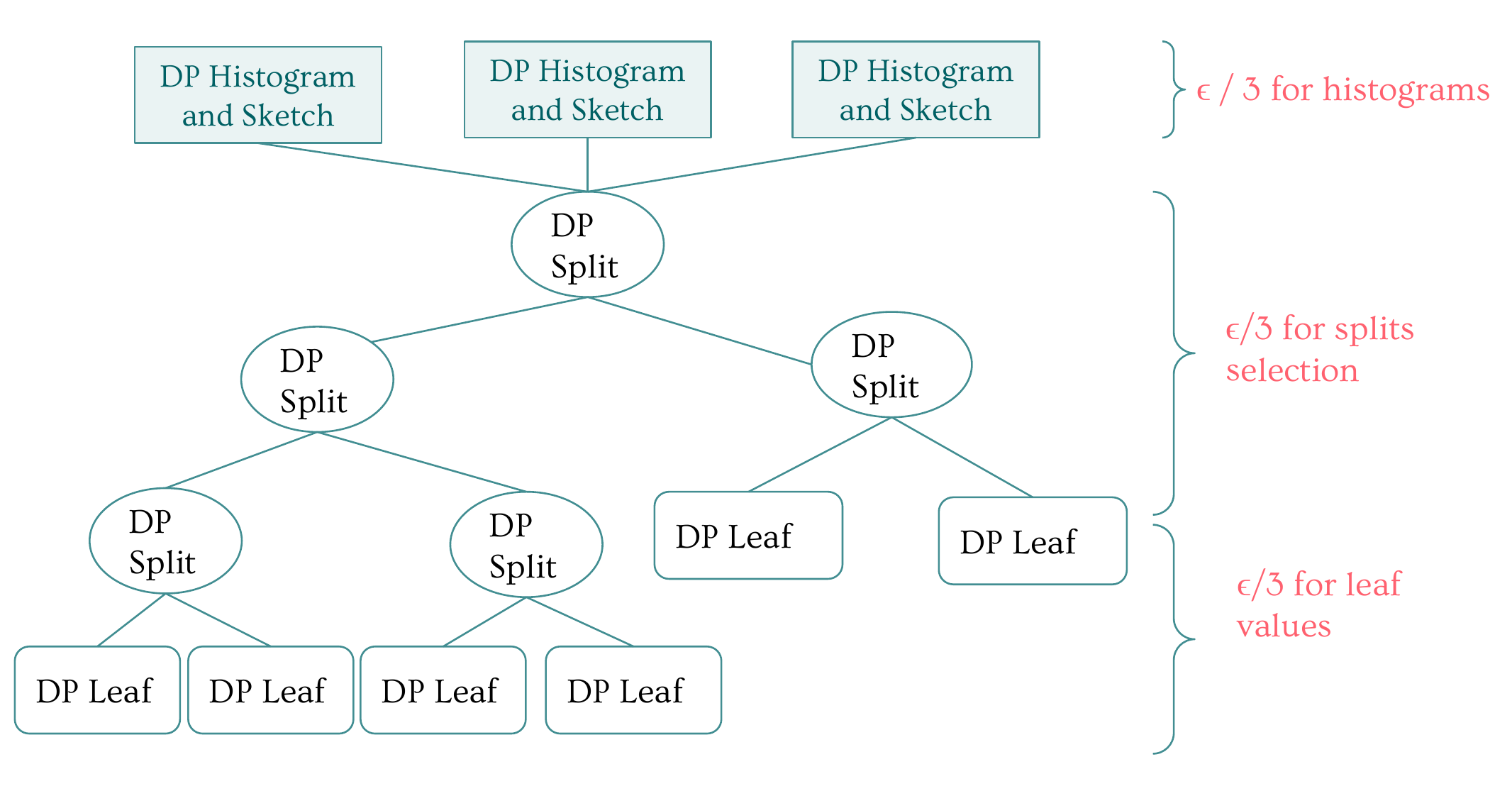}
\label{sketchAccuracy}
\caption{Schema of the DP mechanisms in a single tree. A histogram is built for each feature and perturbated with Laplace mechanisms.
It is then used to build the XGBoost sketches (one per feature).
Then, as in \cite{li2020privacy} candidates splits are then selected from the set of DP candidates with an exponential mechanism.
Finally the leaf values are perturbated with a Laplace mechanism.
Note since the tree is slightly unbalanced, the privacy consumption of the exponential mechanisms as is not optimal.} 
\end{figure}

Please also note that only \emph{basic composition} is used in this work: we only use $\epsilon$-DP mechanisms and compose them by summing up $\epsilon$'s.
If one is ready to accept the slightly relaxed \emph{approximate differential privacy} ($\epsilon,\delta$-DP) with some $\delta>0$ then
composition can be much cheaper in terms of privacy loss, with $\epsilon$ growing as the square root of the number of mechanisms composed sequentially (instead of linearly).

\section{Numerical Experiments}

\label{numExpSection}

We conduct numerical experiments following \cite{li2020privacy}: we use the same datasets,
some are synthetic (one for regression and one for regression), and use other real-world datasets.
We also turn all the problems into regression problems, so that the universal gradient filtering bound is $g_l^* = 1$.
We fix the number of trees to $20$, and use a regularization parameter $\lambda = 0.1$.
The learning rate parameter is set is $0.3$. The minimum number of children to split is set to $500$ for datasets with more than a hundred thousand entries,
and to $50$ for the smaller ones. Finally, row subsampling is applied with a fraction of $0.1$ (each tree is trained 10\% of the dataset).
The following table (figure~\ref{resTable}) summarizes the results obtained with the method described above and a non-DP benchmark.
The errors reported are the average accuracy score in the case of classification problems, and the RMSE for regression, over 5 trials. 

Figure \ref{figError} shows the test accuracy error compared to the method of \cite{li2020privacy}. As pointed out, the algorithm they describe is for the exact learning algorithm (looping over all the splits), which is very slow and unable to handle large datasets, however the implementation they give uses LightGBM ability to build a histogram of feature values, in a way that is very similar to XGBoost. The resulting implementation and numerical results in \cite{li2020privacy} are thus not entirely differentially private. However, as in the case of the XGBoost sketching algorithm, the privacy budget needed to correct this is not large and we think it still makes sense to compare our results with the ones given in their paper. 

The results displayed are very encouraging especially for the higher privacy-regimes ($\epsilon$ = 1). As expected, using subsampling and a minimum weight stopping criterion allows to profit from the natural prevention of overfitting and a significant noise reduction, leading us to obtain almost half the test classification error on the Covtype dataset for $\epsilon = 1$. As in the case of \cite{li2020privacy} we find that the differential-privacy mechanisms do not introduce any significant runtime overhead as compared to the XGBoost baseline, for instance on the covtype dataset, the mean runtime per tree for the DP version is $1$ second compared to a non-DP mean of $0.7$ second.   

\begin{figure}[h]
\begin{tabular}{c|c|c|c|c|c|c|c}
   $\epsilon$ & $1.0$ & $2.0$ & $4.0$ & $6.0$ & $8.0$ & $10.0$ & non-DP benchmark\\ \hline 
   covtype & $22$ & $18$ & $15$ & $17$ & $16$ & $15$ & $11$ \\ \hline 
   adult & $24$ & $18$ & $19$ & $19$ & $18$ & $18$ & $16$ \\ \hline 
   synthetic cls & $12$ & $10$ & $11$ & $8$ & $8$ & $8$ & $7$ \\ \hline 
   synthetic reg (RMSE) & $86.4$ & $76$ & $73$ & $67$ & $61$ & $60$ & $24.1$ \\  \hline 
   abalone (RMSE) & $6$ & $5.5$ & $3.2$ & $2.7$ & $2.6$ & $2.4$ & $2.22$ \\ \hline 
\end{tabular}
\caption{Test error (RMSE or accuracy in \%) on different datasets}
\label{resTable}
\end{figure}

\begin{figure}[h]
\begin{subfigure}{.5\textwidth}
  \centering
  \includegraphics[scale=0.5]{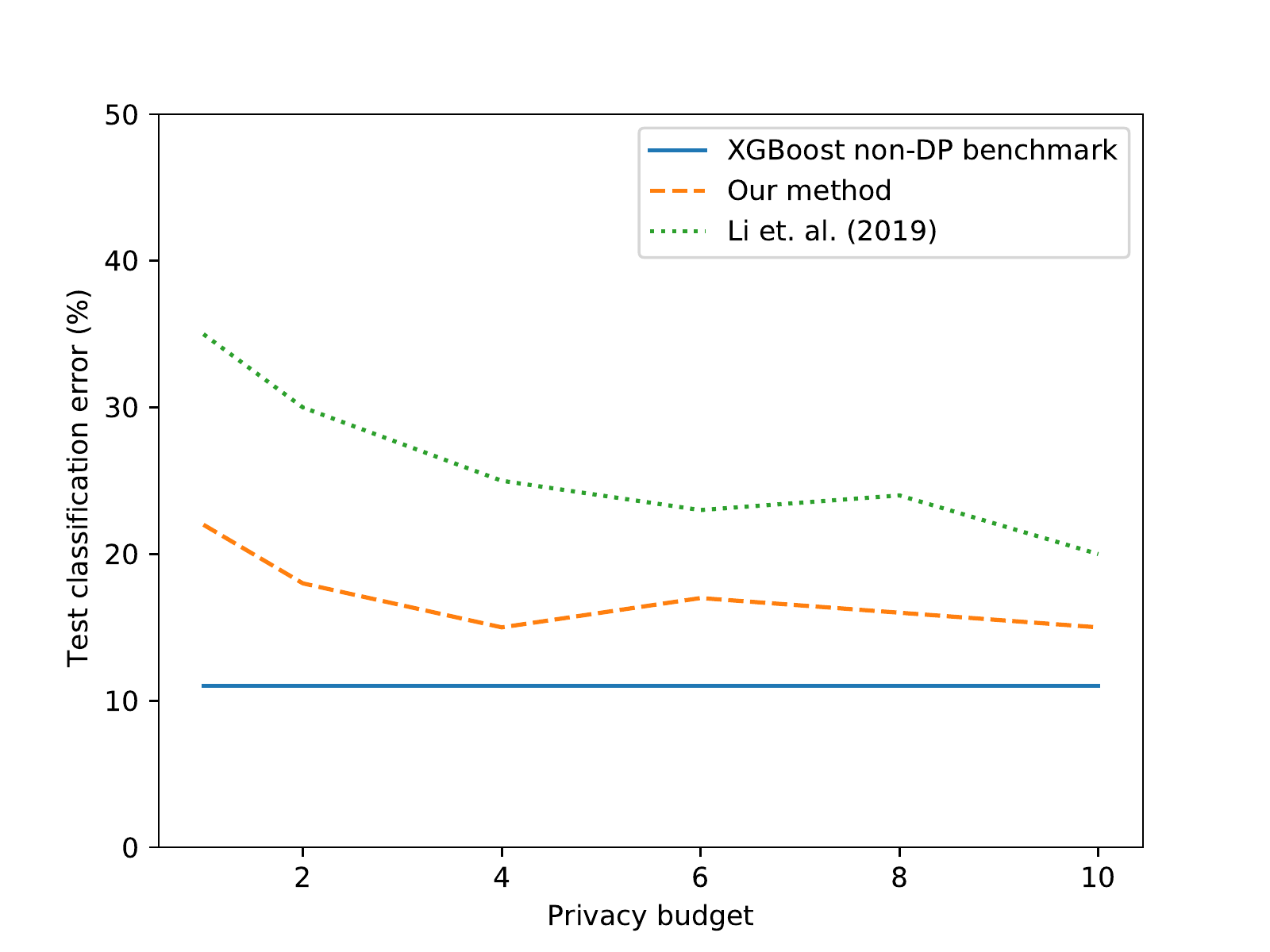}
  \caption{Covtype}
\end{subfigure}
\begin{subfigure}{.5\textwidth}
  \centering
  \includegraphics[scale=0.5]{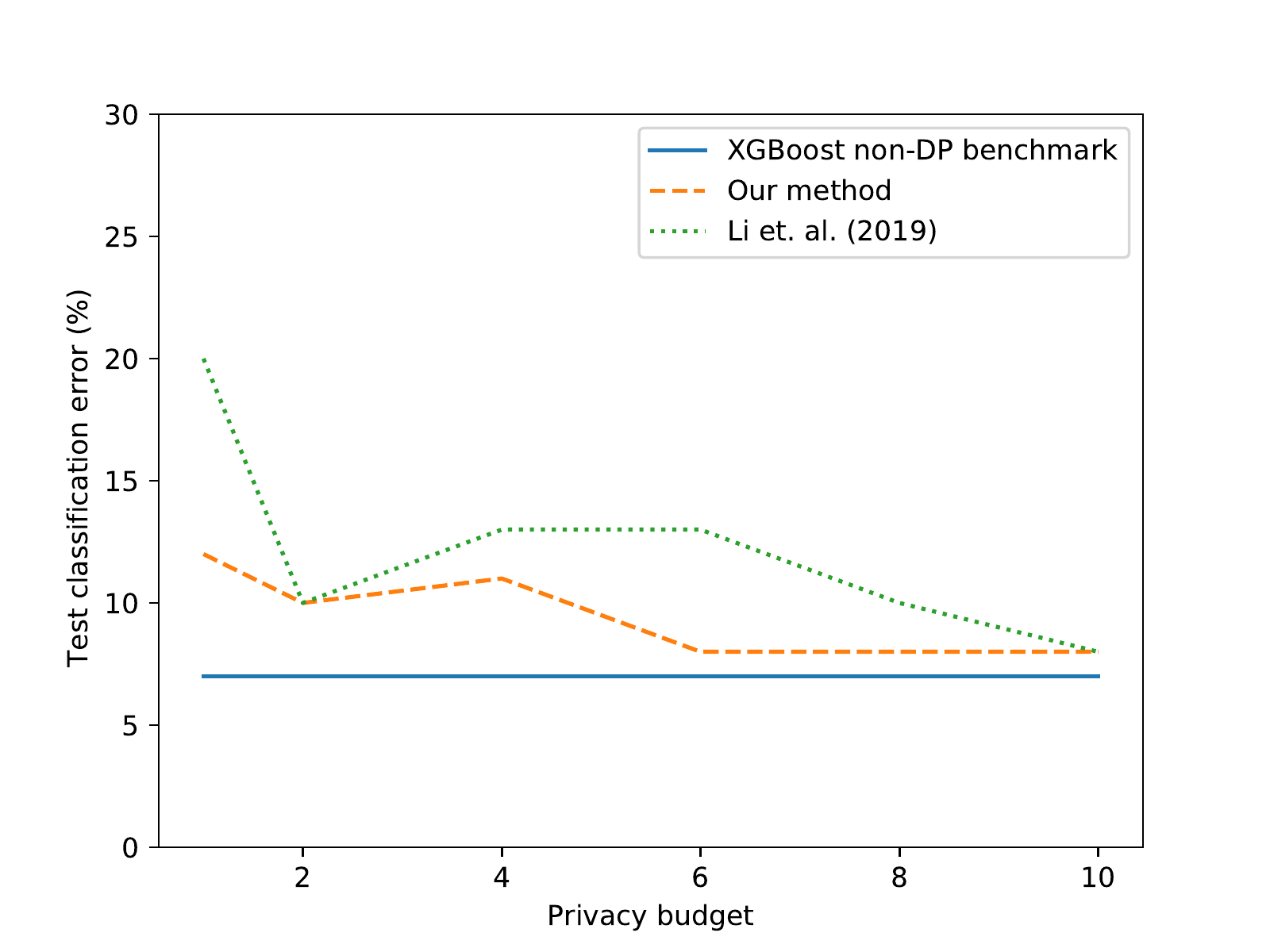}
  \caption{Synthetic cls}
\end{subfigure}
\caption{Test classification error in percentage on a synthetic and real-world (Covtype) classification dataset
as a function of privacy budget $\epsilon$}
\label{figError} 
\end{figure}

\begin{figure}[h]
\begin{subfigure}{.5\textwidth}
  \centering
  \includegraphics[scale=0.5]{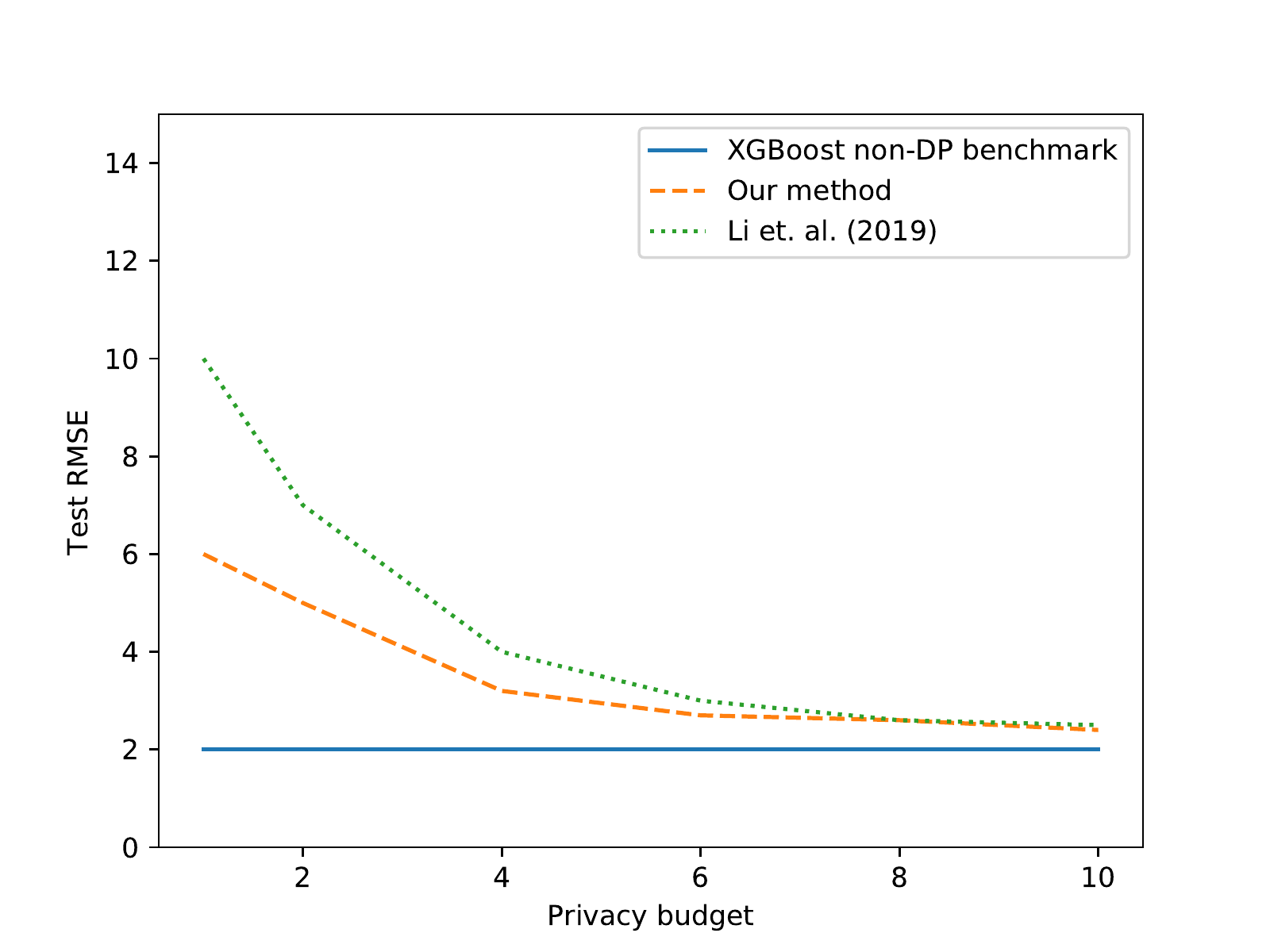}
  \caption{Abalone}
\end{subfigure}
\begin{subfigure}{.5\textwidth}
  \centering
  \includegraphics[scale=0.5]{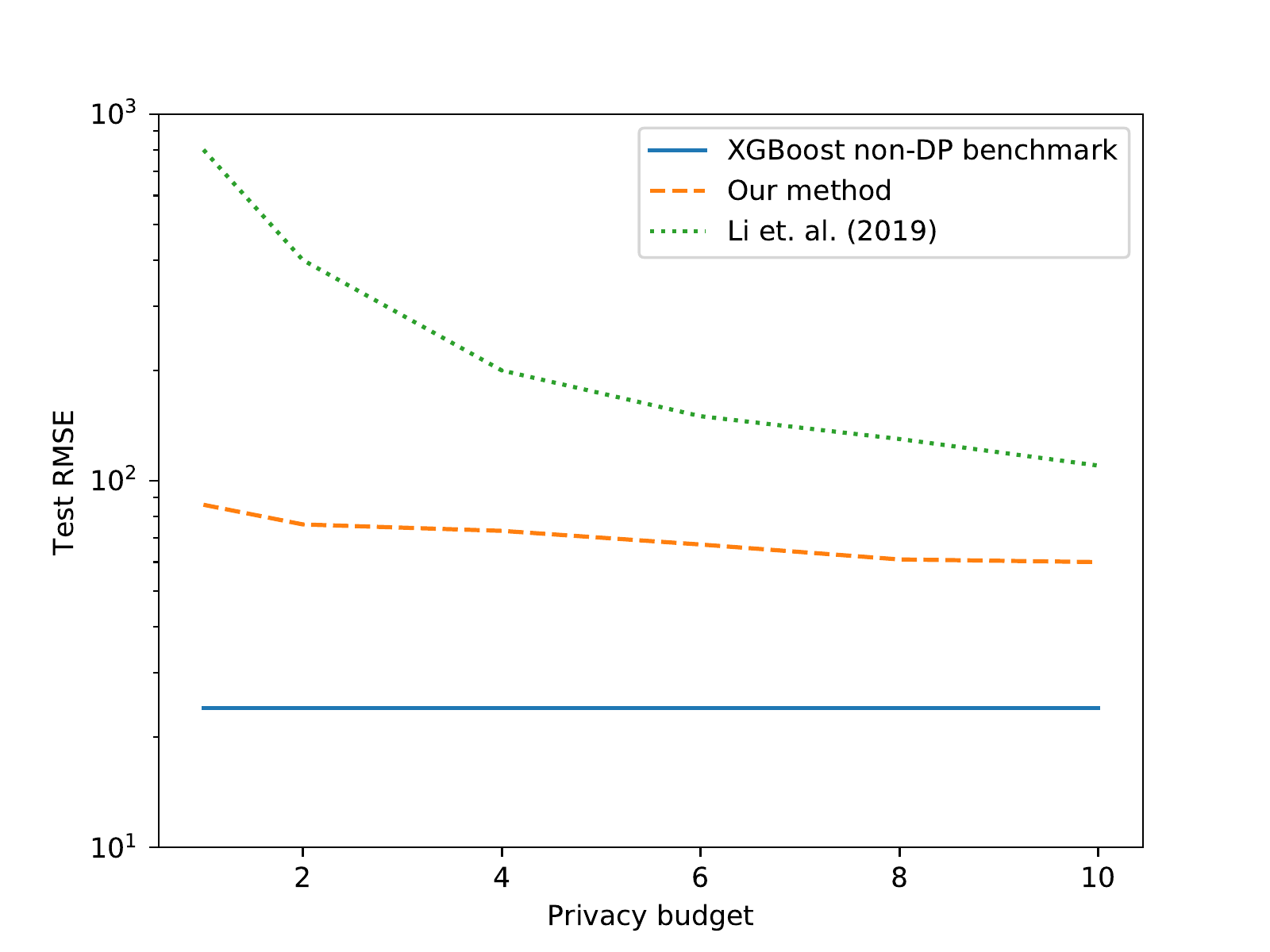}
  \caption{Synthetic reg}
\end{subfigure}
\caption{Test RMSE on a synthetic and real-world (Abalone) regression dataset as a function of 
privacy budget $\epsilon$}
\label{figError} 
\end{figure}

\section{Conclusion} 

We have described a way to implement a fully $\epsilon$ differentially-private gradient boosted trees algorithm based on XGBoost and the work of \cite{li2020privacy}.
We extended their workd to the approximate learning method by building a DP histogram and were able to reduce the crucial noise in the leaf values thanks to widely used hyperparameters that automatically bring differential privacy and often reduce overfitting. By leveraging XGBoost capabilities to handle large-scale datasets and its integration with Spark we thus obtained an efficient method to train DP boosted trees for real-world applications. 

Many improvements remain to be explored. First, a private-by-design quantile sketching algorithm could be studied instead of our simple approach,
perhaps using the state-of-the-art DP quantile algorithms such as JointExp from \cite{gillenwater2021}.
Another topic which could lead to privacy improvement is the distribution of the privacy budget among trees and inside each tree. 

Also, accounting is quite rudimentary in this work. When training large models, approaches such as $f$-DP or Gaussian-DP \cite{dong2019gaussian} could result in
much better privacy consumptions, when \emph{approximate DP} with a small $\delta$ is acceptable.

Beyond accounting, many improvements could be obtained by keeping track of past queries to answers new queries (see \cite{dwork2012privacy}).
Practical implementations of this idea exist \cite{kotsogiannis2019privatesql, kotsogiannis2019architecting} and could be adapted to boosted tree training.
 
\bibliographystyle{alpha}
\bibliography{SarusXGBoost}

\end{document}